\documentclass[10pt,twocolumn,letterpaper]{article}

\usepackage{cvpr}              % To produce the CAMERA-READY version
\usepackage{graphicx}
\usepackage{amsmath}
\usepackage{amssymb}
\usepackage{booktabs}
\usepackage[table,xcdraw]{xcolor}
\usepackage{multirow}
\usepackage{balance}

\makeatletter
\newcommand*\bigcdot{\mathpalette\bigcdot@{.5}}
\newcommand*\bigcdot@[2]{\mathbin{\vcenter{\hbox{\scalebox{#2}{$\m@th#1\bullet$}}}}}
\makeatother

\usepackage[pagebackref,breaklinks,colorlinks]{hyperref}

\usepackage[capitalize]{cleveref}
\crefname{section}{Sec.}{Secs.}
\Crefname{section}{Section}{Sections}
\Crefname{table}{Table}{Tables}
\crefname{table}{Tab.}{Tabs.}

\definecolor{c2}{HTML}{FBD9BD}
\definecolor{c3}{HTML}{fe793d}
\definecolor{c4}{HTML}{eedeb0}
\definecolor{rouse}{rgb}{0.981,0.961,0.941}

\def\cvprPaperID{8933} % *** Enter the CVPR Paper ID here
\def\confName{CVPR}
\def\confYear{2024}

\begin{document}

\title{Wavelet-based Fourier Information Interaction with Frequency Diffusion Adjustment for Underwater Image Restoration}
% 还是叫光照退化图像增强？Low-light Image Enhancement
% 本文旨在解决一类光照退化的问题。在进行颜色恢复，保证全局视觉保真的同时也对细节纹理进行修复

%%%%%%%%% AUTHORS - PLEASE UPDATE
\author{Chen Zhao,
        Weiling Cai$^{}$\thanks{Corresponding Author}\,~,
        Chenyu Dong and Chengwei Hu\\
	School of Artificial Intelligence, Nanjing Normal University
%  {\tt\small\{chunminghe19990224,li.gml.kai,lloong.x,yulun100,cszguo\}@gmail.com} \\ 
% {\tt\small yachaozhang@stu.xmu.edu.cn}, {\tt\small li.xiu@sz.tsinghua.edu.cn} 
 }

\maketitle
% \renewcommand{\thefootnote}{\fnsymbol{footnote}} %将脚注符号设置为fnsymbol类型，即特殊符号表示
% \footnotetext[1]{Corresponding author.} %对应脚注[2]
% \renewcommand{\thefootnote}{\arabic{footnote}}

%%%%%%%%% ABSTRACT
\begin{abstract}
   Underwater images are subject to intricate and diverse degradation, inevitably affecting the effectiveness of underwater visual tasks. However, most approaches primarily operate in the raw pixel space of images, which limits the exploration of the frequency characteristics of underwater images, leading to an inadequate utilization of deep models' representational capabilities in producing high-quality images. 
   In this paper, we introduce a novel Underwater Image Enhancement (UIE) framework, named WF-Diff, designed to fully leverage the characteristics of frequency domain information and diffusion models.
   WF-Diff consists of two detachable networks: Wavelet-based Fourier information interaction network (WFI2-net) and Frequency Residual Diffusion Adjustment Module (FRDAM). With our full exploration of the frequency domain information, WFI2-net aims to achieve preliminary enhancement of frequency information in the wavelet space. Our proposed FRDAM can further refine the high- and low-frequency information of the initial enhanced images, which can be viewed as a plug-and-play universal module to adjust the detail of the underwater images. With the above techniques, our algorithm can show SOTA performance on real-world underwater image datasets, and achieves competitive performance in visual quality. The code  is available at \href{https://github.com/zhihefang/WF-Diff}{here}.
\end{abstract}

%% narrow the gap between equations and sentences
%\setlength{\abovedisplayskip}{2pt}
%\setlength{\belowdisplayskip}{2pt}

% \clearpage
%%%%%%%%% BODY TEXT
%\vspace{-3mm}
\section{Introduction}%\vspace{-1mm}
\label{sec:intro}
Underwater image restoration is a practical but challenging technology in the field of underwater vision, widely used for tasks, such as underwater robotics\cite{McMahon} and underwater object tracking\cite{LangisS}. Due to light refraction, absorption, and scattering in underwater scenes, underwater images are usually severely distorted, with low contrast and blurriness \cite{AkkaynakTSLTI17}. Therefore, clear underwater images play a critical role in fields which need to interact with the underwater environment. The main goal of underwater image enhancement (UIE) is to obtain high-quality images by removing scattering and correcting color distortion in degraded images. UIE is crucial for vision-related underwater tasks.

\begin{figure}[!t]
	\centering
	\includegraphics[width=\linewidth]{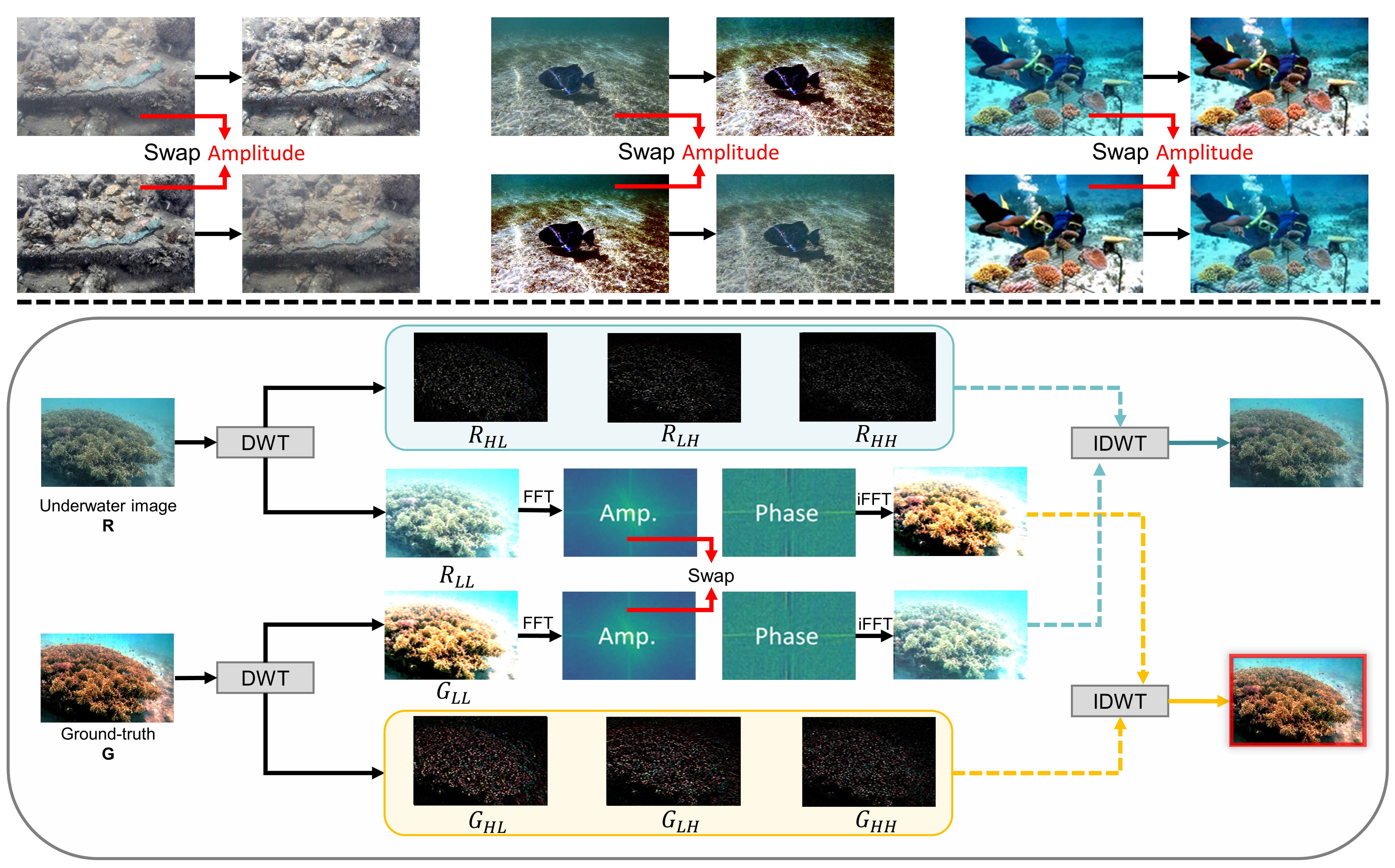}
	%\vspace{0.2cm}
	\caption{Our motivations. The amplitude and
		phase are produced by Fast Fourier Transform (FFT) and the recombined images are obtained by Inverse FFT
		(IFFT). We further explore the frequency properties for underwater images in Wavelet space. 
	}
\end{figure}

\begin{table}[h]
	\centering
	\small
	
	\caption{Evaluation of using different frequency domain transformation strategies on the UIEBD \cite{PengZB23}. S1 refers to swapping the amplitude in original images, S2 refers to only swapping the amplitude of low-frequency sub-images in wavelet space, and S3 refers to swapping the amplitude of low-frequency sub-images and high-frequency sub-images in wavelet space.   }
	%\vspace{0.1cm}
	\scalebox{0.92}{
		\setlength{\tabcolsep}{2mm}{
			\begin{tabular}{c|c|c|c|c}
				\toprule
				\textbf{Strategy} &\textbf{PSNR} $\uparrow$ &\textbf{SSIM} $\uparrow$ &\textbf{LPIPS}$\downarrow$ &\textbf{FID}$\downarrow$\\ 
				% \midrule
				\midrule
				\textbf{S1}&28.68 &0.9027 &0.0957 &28.70  \\
				\textbf{S2} &27.10 &0.8813 &0.1023  &33.04 \\
				\textbf{S3} &\textbf{29.97} &\textbf{0.9343} &\textbf{0.0820} &\textbf{23.55}  \\
				\bottomrule
	\end{tabular}}}
	\vspace{-0.3cm}
	\label{tab:data augmentation}
\end{table}

% Recently, many deep learning based methods\cite{UIEBD,EUVP-FUnieGan,ucolor,cwr} have been proposed to address this problem. 

To address this problem, traditional UIE methods based on the physical properties of the underwater images were proposed  \cite{PengC17,DrewsNMBC13,PengCC18,RavisankarSR18,LiGCPW16}. These methods investigate the physical mechanism of the degradation caused by color cast or scattering and compensate them to enhance the underwater images. 
However, these physics-based model with limited representation capacity cannot address all the complex physical and optical factors underlying the underwater scenes, which leads to poor enhancement results under highly complex and diverse underwater scenes. Recently, some learning based methods \cite{TangKI23,PengZB23,LiGRCHKT20,FabbriIS18} for UIE can produce better results, since neural networks have powerful feature representation and nonlinear mapping capabilities. It can learn the mapping of an image from degenerate to clear from a substantial quantity of paired training data. %Yang et al. \cite{jian2018} proposed a reflected light-aware multi-scale progressive restoration network to obtain images with both color equalization and rich texture in various underwater scenes. %A U-Net with spatial- and channel-wise normalization was employed to handle the variability of underwater scenes \cite{zhenqi2022}. 
%Huang et al. propose a mean teacher based semi-supervised network, which effectively leverages the knowledge from unlabeled data. 
However, most previous methods are based on the raw pixel space of images, with limited exploration of the properties of the frequency space for underwater images, which results in an inability to effectively harness the representation power of the deep models for generating high-quality images.

%Inspired by previous Fourier based works
Building on insights from previous Fourier-based works \cite{abs-2309-04089,HuangLZYZHZX22}, we explore the properties of the Fourier frequency information for UIE task, as illustrated in Figure 1. Given two images (underwater image and its corresponding ground-truth), we swap their amplitude components and combine them with corresponding phase components in the Fourier space. The recombined results show that the visual appearance are swapped following the amplitude swapping, which indicates the degradation information of underwater images is mainly contained in the amplitude component. We further explore the properties of the amplitude components in Wavelet space. Specifically, the images can be decomposed into low-frequency sub-images and high-frequency sub-images using discrete wavelet transformations (DWT), and then we swap amplitude components of low-frequency sub-images. From visual results, we can find a similar phenomenon, which means the color degradation information is mainly contained in low-frequency sub-images, and the texture and detail degradation information is mainly contained in high-frequency sub-images. Table 1 shows the quantitative evaluation of the different frequency domain strategies, proving that our discovery is objective. %From visual results, we can find a similar phenomenon, which means the color degradation information of underwater images is mainly contained in low-frequency sub-images, and the texture and detail degradation information is mainly contained in high-frequency sub-images. %\textbf{Consequently, how to adequately exploit the properties frequency domain information and effectively incorporate them into a unified image enhancement network is an open issue.}
Consequently, how to adequately exploit the properties of frequency domain information and effectively incorporate them into a unified image enhancement network is an crucial issue.

Recently, diffusion-based methods \cite{DDPM,DDIM} have garnered significant attention due to their outstanding performance in image synthesis \cite{RombachBLEO22,SahariaCSLWDGLA22} and restoration tasks \cite{ChoiKJGY21,WangYZ23, abs-2308-13164}. These methods rely on a hierarchical denoising autoencoder architecture, enabling them to iteratively reverse a diffusion
process and achieve high-quality mapping from randomly
sampled Gaussian noise to target images or latent distributions \cite{DDPM}. Tang et al. \cite{TangKI23} present an image enhancement approach with diffusion model in underwater scenes. While standard diffusion models exhibit sufficient capability, unforeseen artifacts may arise as a result of the diversity introduced during the sampling process from randomly generated Gaussian noise to images \cite{Doc}. %Although standard diffusion models are capable enough, unexpected artifacts occur due to the diversity of the sampling process from randomly sampled Gaussian noise to images \cite{Doc}. %Although standard diffusion models are capable enough, there exist several challenges for UIE tasks. As shown in Fig. 2, the DM-underwater \cite{TangKI23} generate images with color distortion or artifacts. This occurs because the reverse process starts from randomly sampled Gaussian noise to image, which can lead to unexpected artifacts due to the diversity of the sampling process \cite{Doc}.  %since the reverse process starts from a random sampled Gaussian noise, resulting in the final result may have unexpected chaotic content due to the diversity of the sampling process. %despite conditional inputs can be used to constrain the output distribution. 
Furthermore, the diffusion model needs to recover both the high and low-frequency information of images, which limits their ability to focus on fine-grained information, missing out on texture and details. %inhibits the model from focusing on the fine-grained information, potentially missing texture details. 
Thereby, it is very crucial that the powerful representation  capabilities of diffusion models can be fully utilized. 

%This motivation inspires us to design a two-branch network to recover the texture and lightness information of underwater images respectively in wavelet space, which can increase the model's focus on fine-grained information to enhance the texture detail of the images.
\begin{figure*}[t]
	\centering
	\includegraphics[width=0.98\linewidth]{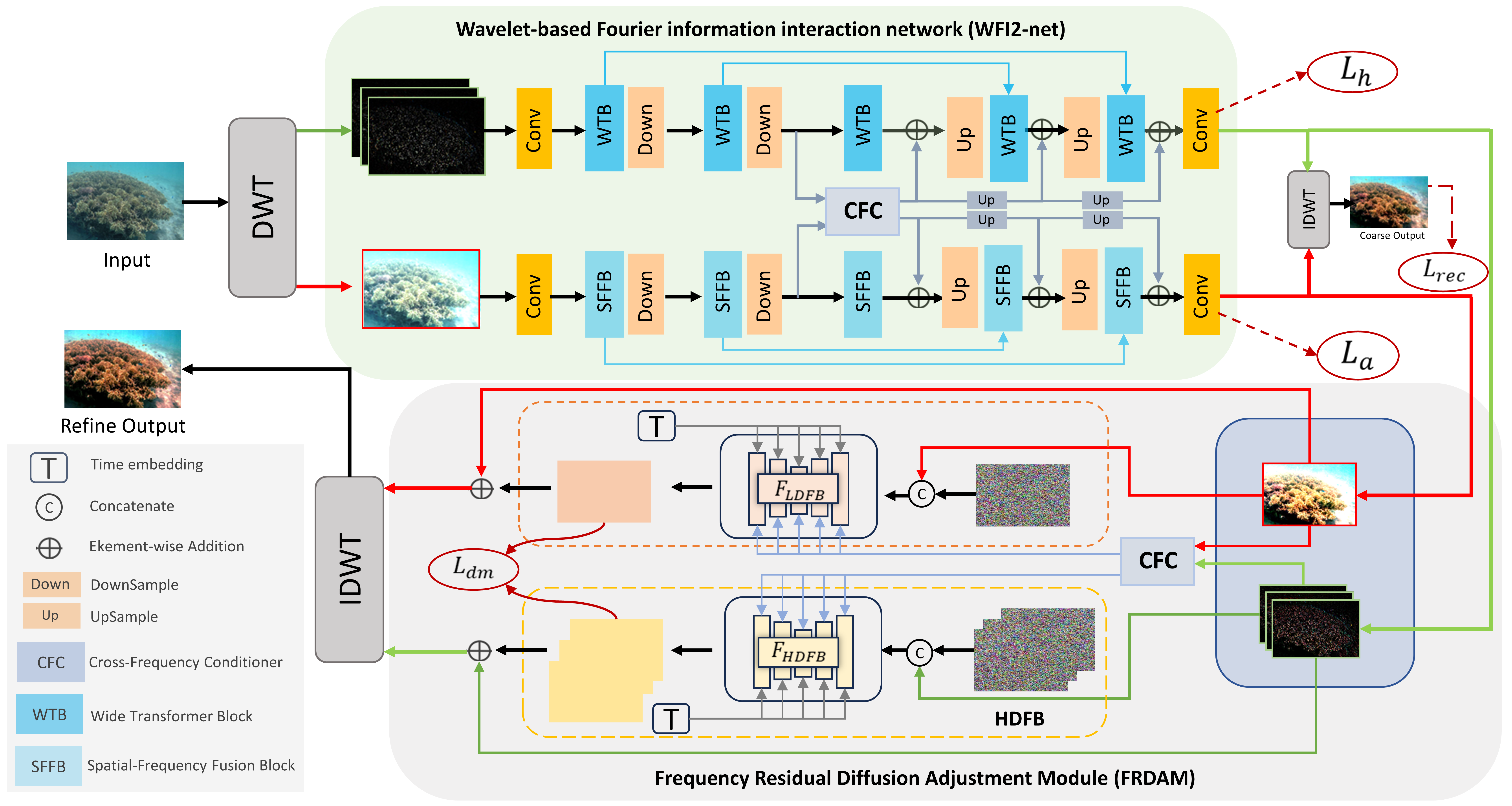}
	%\vspace{0.3cm}
	\caption{Overall framework of WF-Diff. It contains two detachable networks, Wavelet-based Fourier information interaction network (WFI2-net) and Frequency Residual Diffusion Adjustment Module (FRDAM). FRDAM consists of low-frequency diffusion branch (LDFB) and high-frequency diffusion branch (HDFB), which aims to further adjust the high- and low-frequency information of the initial enhanced images. Furthermore, the proposed cross-frequency conditioner (CFC) aims to achieve the cross-frequency interaction of high- and low-frequency information.} 
	\vspace{0.1cm}
	\label{fig:2}
\end{figure*}

In this paper, we develop a novel UIE framework to fully exploit the properties of frequency domain information and diffusion models, called WF-Diff, which mainly consists of two stages: frequency preliminary enhancement and frequency diffusion adjustment. The first stage aims to preliminarily enhance the high-frequency and low-frequency components of underwater images by utilizing the frequency domain characteristics. Specifically, we first convert the input images into the wavelet space using discrete wavelet transformations (DWT), obtaining an average coefficient that represents the low-frequency content information of the input image and three high-frequency coefficients that represent sparse vertical, horizontal, and diagonal details of the input image. Then, we design a wavelet-based fourier information interaction network (WFI2-net), fullly integrating the characteristics of Transformer \cite{LiuL00W0LG21} and Fourier prior information to enhance high- and low-frequency content, respectively. Moreover, to achieve interaction of high- and low-frequency information, we propose a cross-frequency conditioner (CFC) to further improve the generation quality. The target of the second stage is to make adjustment to the initial enhanced coarse results in terms of details and textures via diffusion model. Consequently, we propose a frequency residual diffusion adjustment module (FRDAM). Unlike previous diffusion-based work, FRDAM learns the residual distribution of  high- and low-frequency information between the ground-truth and the initial enhanced results using two diffusion models in the wavelet space, which can not only increase the model's focus on fine-grained information but also mitigate the adverse effects of the diversity of the sampling process.

In summary, the main contributions of our method are as follows:

%wavelet-based fourier information interaction network (WFIInet) and the frequency residual diffusion adjustment module (FRDAM).
\begin{itemize}
	\item We explore in depth the properties of the frequency domain for underwater images. Based on the properties and diffusion model, we propose a novel UIE framework, named WF-Diff, with the goal of achieving frequency enhancement and diffusion adjustment.
	
	\item We propose a frequency residual diffusion adjustment module (FRDAM) to further refine the high- and low-frequency information of the initial enhanced images. FRDAM can be viewed as a plug-and-play universal module to adjust the detail of the underwater images.
	
	\item We propose a cross-frequency conditioner (CFC) to achieve the cross-frequency interaction of high- and low-frequency information.
	
	\item Experimental results compared with SOTAs considerably show
	that our developed WF-Diff performs the superiority against
	previous UIE approaches, and extensive ablation experiments can demonstrate the effectiveness of our contributions.
	
\end{itemize}

\section{Related Works} \label{sec:related_work}

\subsection{Underwater Image Enhancement }
Currently, existing UID methods can be briefly categorized into
the physical and deep model-based approaches \cite{PengC17,DrewsNMBC13,PengCC18,TangKI23,PengZB23,LiGRCHKT20}. Most UID methods based on the physical model utilize prior knowledge to establish models, such as water dark channel priors \cite{PengC17}, attenuation curve priors \cite{WangLC18}, fuzzy priors \cite{ChiangC12}. In addition, Akkaynak and Treibitz \cite{AkkaynakT19} proposed a method based on the revised physical imaging model. However, the depth map of the underwater scene is difficult to obtain. This leads to unstable performance, which usually suffers from severe color cast and artifacts. Therefore, the manually established priors restrain the model’s robustness and scalability under the complicated and varied circumstances.  Recently, deep learning-based methods \cite{TangKI23,PengZB23,LiGRCHKT20} achieve acceptable performance. To alleviate the need for real-world underwater paired training data, many methods introduce GAN-based framework for UIE \cite{LiSEJ18,IslamXS20,FabbriIS18}, such as WaterGAN \cite{LiSEJ18}, UGAN \cite{FabbriIS18} and UIE-DAL \cite{UplavikarWW19}. Recently, some complex frameworks are proposed and achieve the-state-of-the-art performance \cite{PengC17,DrewsNMBC13}. Ucolor \cite{LiAHCGR21} combined the underwater physical imaging model in the raw space and designed a medium transmission guided model. Yang et al. \cite{angLL23} proposed a reflected light-aware multi-scale progressive restoration network to obtain images with both color equalization and rich texture in various underwater scenes.
Huang et al. \cite{HuangWL0L23} proposed a mean teacher based semi-supervised network, which effectively leverages the knowledge from unlabeled data. However, most previous methods are based on the spatial domain, with limited exploration of the frequency space for underwater images, which results in an inability to effectively harness the representation power of the deep models.

\subsection{Diffusion Model}
Recently, Diffusion Probabilistic Models (DPMs) \cite{DDPM,DDIM} have been
widely adopted for conditional image generation  \cite{ChoiKJGY21,WangYZ23, abs-2308-13164,WhangDTSDM22}. Saharia et al.  \cite{SahariaCCLHSF022} proposes Palette, which has demonstrated the excellent performance of diffusion models in the field of conditional image generation, including colorization, in-
painting and JPEG restoration. Tang et al. \cite{TangKI23} presented an image enhancement approach with diffusion model in underwater scenes.  However, the reverse process starts from randomly sampled Gaussian noise to full images  \cite{Doc}, which can lead to unexpected artifacts due to the diversity of the sampling process. Furthermore, the diffusion model needs to recover both the high and low-frequency information in images, which limits their ability to focus on fine-grained information. Consequently, how to incorporate diffusion models into a unified underwater image enhancement network is a vital issue.

\section{Methodology}
%WF-Diff aims to fully exploit the properties of frequency domain information and diffusion models.
\subsection{Overall Framework}
Given an underwater image as input, our goal is to learn a network to generate an output that eliminates the color cast from input while enhancing the image details. The overall framework of WF-Diff is shown in Figure 2. WF-Diff is designed to fully leverage the characteristics of frequency domain information and powerful ability of diffusion models. Specifically, WF-Diff consists of two detachable networks: Wavelet-based Fourier information interaction network (WFI2-net) and Frequency Residual Diffusion Adjustment Module (FRDAM). We first convert the input into the wavelet space using discrete wavelet transformations (DWT), obtaining an low-frequency coefficient and three high-frequency coefficients. WFI2-net  is dedicated to achieving preliminary enhancement of frequency information. We fullly integrate the characteristics of Transformer and Fourier prior Information, and design wide transformer block (WTB) and spatial-frequency fusion block (SFFB) to enhance high- and low-frequency content respectively. FRDAM consists of low-frequency diffusion branch (LDFB) and high-frequency diffusion branch (HDFB), which aims to further adjust the high- and low-frequency  information of the initial enhanced images. Note that, our proposed FRDAM learns the residual distribution of  high- and low-frequency information between the ground-truth and the initial enhanced results using two diffusion models, respectively. Additionally, the proposed cross-frequency conditioner (CFC) strives to achieve cross-frequency interaction between high- and low-frequency information.
\begin{figure*}[t]
	\centering
	
	\includegraphics[width=0.98\linewidth]{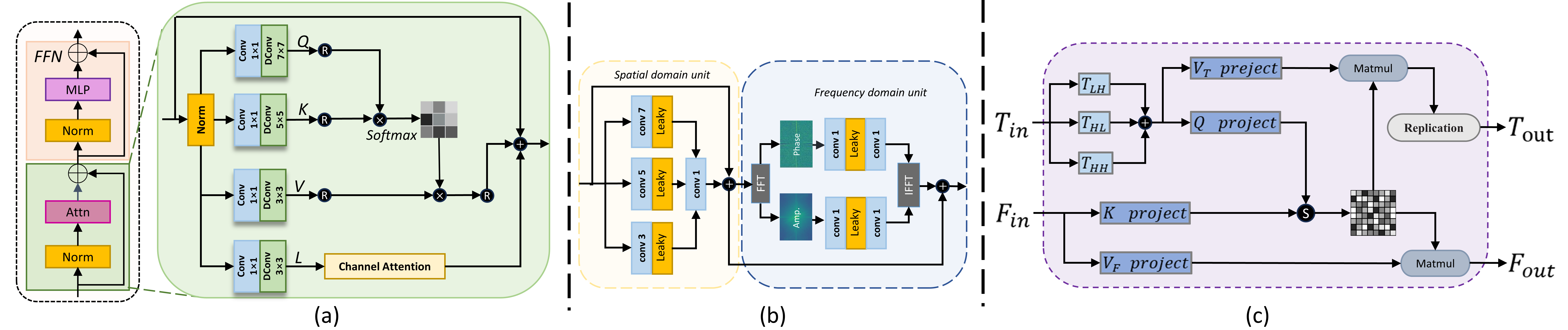}
	%\vspace{0.3cm}
	\caption{The detailed architecture of (a) Wide Transformer Block, (b) Spatial-Frequency Fusion Block and (c) Cross-Frequency Conditioner.} 
	\label{fig:2}
\end{figure*}
\subsection{ Discrete Wavelet and Fourier Transform}

Discrete Wavelet transform (DWT) has been widely applied to low-level vision tasks \cite{HuangHST17,KangCYY18}. %such as image super-resolution [16] and denoising [17]. %Traditional physics-based methods have used DWT to decompose the underwater images to improve the contrast and resolution. Unlike previous methods, We aim to integrate the advantages of the frequency domain and the diffusion model.  
We firstly use DWT to decompose an input into multiple frequency sub-bands so that we can achieve the color correction of low-frequency information and detail enhancement of high-frequency information, respectively. Given a underwater image as input $I\in\mathbb{R}^{H\times W\times c}$, we use DWT with Haar wavelets to decompose the input. Haar wavelets consist of the low-pass filter $L$, and the high-pass filter $H$, as follows:
\begin{equation}L=\frac1{\sqrt{2}}[1,1]^T,H=\frac1{\sqrt{2}}[1,-1]^T.\end{equation}

We can obtain four sub-bands, which can be expressed as:

\begin{equation}
	I_{LL},\{I_{LH},I_{HL},I_{HH}\}=\text{DWT}(I),
\end{equation}
where $I_{LL},\{I_{LH},I_{HL},I_{HH} \} \in \mathbb{R}^{\frac H2\times\frac W2\times c}$ represent the low-frequency component of the input and high-frequency components in the vertical, horizontal, and diagonal directions, respectively.  More specifically, the low-frequency component contains
the content and color information of the input image, and
the other three high-frequency coefficients contain details information of global structures and textures \cite{RavisankarSR18}. %Consequently, we utilize advantage of transformer modeling global information to enhance high-frequency coefficients. 
The sub-bands are downsampled to half-resolution of the input but do not result in information loss due to the biorthogonal property of DWT. For low-frequency component $I_{LL}$, we will explore its properties in Fourier space. 

Then, we introduce the operation of the Fourier transform \cite{abs-2309-04089}. Given a image $x\in\mathbb{R}^{H\times W\times 1}$, whose shape is $H\times W$, the Fourier transform $\mathcal{F}$ which converts $x$ to the Fourier space $X$ can be expressed as:
\begin{equation}
	\scriptsize
	\mathcal{F}\left(x\right)(u,v)=X(u,v)=
	\frac{1}{\sqrt{HW}}\sum_{h=0}^{H-1}\sum_{w=0}^{W-1}x(h,w)e^{-j2\pi(\frac{h}{H}u+\frac{w}{W}v}),
\end{equation}
where $h, w$ are the coordinates in the spatial space and $u, v$ are the
coordinates in the Fourier space. $\mathcal{F}^{-1}$ denotes the inverse transform of $\mathcal{F}$. Complex component $X(u,v)$ in the Fourier space can be represented
by a amplitude component $\mathcal{A}(X(u,v))$ and a phase component $\mathcal{P}(X(u,v))$ as follows:
\begin{equation}
	\begin{array}{l}\mathcal{A}(X(u,v))=\sqrt{R^2(X(u,v))+I^2(X(u,v))},\\ \mathcal{P}(X(u,v))=arctan[\frac{I(X(u,v))}{R(X(u,v))}],\end{array}
\end{equation}
where $R(x)$ and $I(x)$ represent the real and imaginary parts of $X(u, v)$, respectively. Note that, the Fourier operation can be computed alone in each channel for feature maps.

According to Figure 1 and Table 1 (our motivation), we conclude that the color degradation information of underwater images is mainly contained in the amplitude component of low-frequency sub-band, and the texture and detail degradation information is mainly contained in high-frequency sub-bands.

\subsection{ Frequency Preliminary Enhancement}

Based on the above analysis, in frequency preliminary enhancement stage, we design a simple but effective WFI2-net with an parallel encoder-decoder (U-Net-like) format to restore the amplitude component of low-frequency information and high-frequency components, respectively. We also utilize
skip connections to connect the features at the same level in
the encoder and decoder. For high-frequency branch, we utilize advantage of transformer modeling global information to enhance high-frequency coefficients. We design wide transformer block (WTB) using multi-scale information, aiming to model long range dependencies. Our low-frequency branch aims to restore the amplitude component in Fourier space. In order to obtain rich frequency and spatial information, we design spatial-frequency fusion block (SFFB).

\noindent\textbf{Wide Transformer Block.} WTB is shown in Figure 3 (a). Given $I_{LH},I_{HL},I_{HH} \in \mathbb{R}^{\frac H2\times\frac W2\times c}$, WTB firstly obtains their embedding features $T_{in} \in \mathbb{R}^{3\times \frac H2\times\frac W2\times C}$
through convolution projection. To be specific, WTB  are composed of an attention (Atten) module and
a feed-forward network (FFN) module, and the computation can be denoted in the WTB as:
\begin{equation}\hat{T}_i=SA(Q,K,V)+CA(L)+T_{i-1},\end{equation}
\begin{equation}Q,K,V,L=\text{Split}\Big(W_dW_p\big(Norm(T_{i-1}))),\end{equation}
\begin{equation}T_i=FFN(Norm(\hat{T}_i))+\hat{T}_i,\end{equation}
where  $SA$ and $CA$ refer to self-attention and channel attention, respectively.  $Norm$ refers to normalization. $T_{i-1}$ represents the input embeddings of the current WTB. $W_d$ and $W_p$ denote 1×1 point-wise convolution and multi-scale kernel depth-wise convolution, respectively; $Split$ refers to the split
operation. $L$ aims to focus on local information. %which aims to reduce computational effort.

\noindent\textbf{Spatial-Frequency Fusion Block.} We show the structure of SFFB in Figure 3 (b), which has a spatial domain unit (SDU) and a frequency domain unit (FDU) for interaction of dual domain representations. In spatial domain unit, we employ multi-scale convolution kernels in order to enlarge the limited spatial receptive field. After obtaining the spatial embeddings $F_{s}$, we firstly utilize the FFT to obtain the amplitude $\mathcal{A}(F_{s})$ and phase $\mathcal{P}(F_{s})$ components. Then, the $\mathcal{A}(F_{s})$ and $\mathcal{P}(F_{s})$ are fed into two layers 1*1 conv to obtain $\mathcal{A}^{^{\prime}}(F_{s} )$ and $\mathcal{P}^{^{\prime}}(F_{s})$. Finally, we
use the IFFT algorithm to map $\mathcal{A}^{^{\prime}}(F_{s} )$ and $\mathcal{P}^{^{\prime}}(F_{s})$ to image space and obtain the frequency embeddings $F_{f}$. The fusion embeddings of spatial domain and frequency domain can be expressed as:
\begin{equation}F_{sf}=F_s+F_f.\end{equation}

\noindent\textbf{Loss Function.} We denote ${I}^{^{\prime}}_{LL}$ as output of low-frequency branch, and ${I}^{^{\prime}}_{LH}, {I}^{^{\prime}}_{HL}$ and ${I}^{^{\prime}}_{HH}$ as output of high-frequency branch. Ground-truth (G) can be decomposed $G_{LL},G_{LH},G_{HL},G_{HH}$ by DWT. The high-frequency loss can be expressd as:
\begin{equation}\mathcal{L}_h=\left\|{I}^{^{\prime}}_{(i)}-G_{(i)}\right\|_2,\end{equation}
where $\begin{aligned}i\in\{LH,HL,HH\}\end{aligned}$. For low-frequency information, we only constrain the amplitude components. Consequently, the low-frequency loss can be expressd as:
\begin{equation}
	\mathcal{L}_a=||\mathcal{A}({I}^{^{\prime}}_{LL} ))-\mathcal{A}(G_{LL})||_1,
\end{equation}
where $\mathcal{A}()$ refers to the amplitude component in Fourier transform. Finally, We further use an adversarial loss in Wasserstein GAN as reconstruction Loss $\mathcal{L}_{rec}$.
\subsection{ Cross-Frequency Conditioner}
The detailed structure of CFC is shown in Figure 3 (c). CFC aims to achieve the cross-frequency interaction. We denote $T_{in}$ and $F_{in}$ as the input features of CFC, which represent high- and low-frequency embeddings.  For the high-frequency embedding features $T_{in} \in \mathbb{R}^{3\times \frac H2\times\frac W2\times C}$, we can obtain $T_{LH},T_{HL},T_{HH} \in \mathbb{R}^{\frac H2\times\frac W2\times c}$ via split operation. By adding these extracted coefficients, we obtain the aggregated high-frequency embeddings. We use different linear projections to construct $Q$ and $K$ in CFC:
\begin{equation}
	Q=Conv_{1\times1}(T_{LH}+T_{HL}+T_{HH}),
\end{equation}
\begin{equation}
	K=Conv_{1\times1}(F_{in}).
\end{equation}

Similarly, $V_{T}$ of high-frequency embeddings and $V_{F}$ of low-frequency embeddings can be obtained:
\begin{equation}
	V_{T}=Conv_{1\times1}(T_{LH}+T_{HL}+T_{HH}),
\end{equation}
\begin{equation}
	V_{F}=Conv_{1\times1}(F_{in}).
\end{equation}

The output feature map $T_{out}$ and $F_{out}$ can then be obtained from the formula:
\begin{equation}T_{out}=R(Softmax(\frac{QK^T}{\sqrt{d_k}})V_{T}),\end{equation}
\begin{equation}F_{out}=Softmax(\frac{QK^T}{\sqrt{d_k}})V_{F},\end{equation}
where $R$ denotes a replication operation, and $\sqrt{d_k}$ is the number of columns of matrix $Q$.
\subsection{ Frequency Diffusion Adjustment}
FRDAM aims to further adjust the high- and low-frequency information using the powerful representation of diffusion model.  Generally, FRDAM can be divided into two
branch, namely low-frequency diffusion branch (LDFB) and high-frequency diffusion branch (HDFB). We adopt the diffusion process proposed in DDPM  \cite{DDPM} to construct the residual distribution of high- and low-frequency information for each branch, which can be described as a forward diffusion process and a reverse diffusion process.

\noindent\textbf{Forward Diffusion Process.} The forward diffusion process can be viewed as a Markov chain progressively adding
Gaussian noise to the data. Given the initial enhanced frequency components ${I}^{^{\prime}}_{i}$ and its ground truth $G_{i}$, $\begin{aligned}i\in\{LL,LH,HL,HH\}\end{aligned}$, we calculate their residual distribution $x_0=G_{i}-{I}^{^{\prime}}_{i}$, then introduce Gaussian
noise based on the time step, as follows:  %${I}_{0}=G_{i}-{I}^{^{\prime}}_{i}$. Thus 

\begin{equation}q(x_t|x_{t-1})=\mathcal{N}(x_t;\sqrt{1-\beta_t}x_{t-1},\beta_tI),\end{equation}
where $\beta_{t}$ is a variable controlling the variance of the noise. Introducing $\alpha_{t}=1-\beta_{t}$, this process can be described as:
\begin{equation}x_t=\sqrt{\alpha_t}x_{t-1}+\sqrt{1-\alpha_t}\epsilon_{t-1},\quad\epsilon_{t-1}\sim\mathcal{N}(0,\mathcal{Z}).\text{(}\end{equation}

With Gaussian distributions are merged, We can obtain :
\begin{equation}q(x_t|x_0)=\mathcal{N}(x_t;\sqrt{\bar{\alpha_t}}x_0,(1-\bar{\alpha_t})I).\end{equation}

\noindent\textbf{Reverse Diffusion Process.} The reverse diffusion process aims to restore the residual distribution from the Gaussian noise. The reverse diffusion can be expressed as:
\begin{equation}p_\theta(x_{t-1}|x_t,x_c^{(l)})=\mathcal{N}(x_{t-1};\mu_\theta(x_t,x_c^{(l)},t),\sigma_t^2\mathcal{Z}),\end{equation}
where we take the LDFB as an example, and $x_{c}^{(l)}$ refers to the conditional image ${I}^{^{\prime}}_{LL}$. $\boldsymbol{\mu}_{\theta}(x_{t},x_{c}^{(l)},t)$ and $\sigma_t^2$ are the mean and variance from the estimate of
step t, respectively. In LDFB and HDFB, we follow the
setup of \cite{DDIM}, they can be expressed as:
\begin{equation}\mu_\theta(x_t,x_c^{(l)},t)=\frac1{\sqrt{\alpha}_t}(x_t-\frac{\beta_t}{(1-\overline{\alpha}_t)}\epsilon_\theta(x_t,x_c^{(l)},t)),\end{equation}

\begin{equation}\sigma_t^2=\frac{1-\overline{\alpha}_{t-1}}{1-\overline{\alpha}_t}\beta_t,\end{equation}
where $\epsilon_\theta(x_t,x_c^{(l)},t)$ is the estimated value with a Unet.

We optimize an objective function for the noise estimated
by the network and the noise $\epsilon^{(l)}$ actually added in LDFB. Therefore,
the diffusion loss process is:
\begin{equation}L_{dm}(\theta)=\|\epsilon^{(l)}-\epsilon_\theta(\sqrt{\overline{\alpha}_t}x_0+\sqrt{1-\overline{\alpha}_t}\epsilon^{(l)},x_c^{(l)},t)\|.\end{equation}

Generally, the frequency diffusion adjustment
process is to refine the high- and low-frequency component of the initial enhancement.  The whole diffusion process can be formulated:
\begin{equation}\hat{I}_{(i)}=\mathcal{F}_{HDFB}(\epsilon_s^{(h)},I^{^{\prime}}_{(i)}), i\in\{LH,HL,HH\},\end{equation}

\begin{equation}\hat{I}_{LL}=\mathcal{F}_{LDFB}(\epsilon_s^{(l)},I^{^{\prime}}_{LL}),\end{equation}
where $\epsilon_s^{(h)}\in\mathbb{R}^{3\times \frac H2\times \frac W2 \times 3}$ and $\epsilon_s^{(l)}\in\mathbb{R}^{\frac H2\times \frac W2 \times 3}$ are Gaussian noise.

Ultimately, the refined frequency components are obtained as the addition of the diffusion generative residual distribution and the initial enhanced frequency components. Then, we employ IDWT to obtain the final generated image:
\begin{equation}I_{final}= IDWT(I^{^{\prime}}_{(i)}+\hat{I}_{(i)}, I^{^{\prime}}_{LL}+\hat{I}_{LL}), i\in\{LH,HL,HH\}\end{equation}

\section{Experiments}

\begingroup
\setlength{\tabcolsep}{6pt}{
	\renewcommand{\arraystretch}{1.2} 
	\begin{table*}[t]
		\centering
		\caption{\label{tab:cmp} Quantitative  comparison of different UIE methods on the UIEBD, LSUI and U45 datasets. The best results are highlighted in bold and the second best results are underlined.}
		%\vspace{0.05in}
		%\vspace{5pt}
		
		\begin{tabular}{l|c|cccccccc|c}
			\toprule
			
			&Methods &UIEWD  & UWCNN  & UIECˆ2-Net  & Water-Net  &  SCNet  & U-color & U-shape  &DM-water & Ours \\
			\hline
			\multirow{4}{*}{\rotatebox{90}{UIEBD}} &FID$\downarrow$ & 85.12  & 94.44 & 35.06 & 37.48 & 33.66 & 38.25 & 46.11 & \underline{31.07} & \textbf{27.85}\\
			&LPIPS$\downarrow$ & 0.3956   & 0.3525 & 0.2033 & 0.2116 &  0.2497 & 0.2337 & 0.2264 & \underline{0.1436} & \textbf{0.1248}\\
			&PSNR$\uparrow$  & 14.65  & 15.40 & 20.14 & 19.35 & 20.41 & 20.71 & 21.25 & \underline{21.88} & \textbf{23.86}\\
			&SSIM$\uparrow$	 & 0.7265 &0.7749 &0.8215 &0.8321 & 0.8235 &0.8411 & \underline{0.8453}&0.8194 &\textbf{0.8730}\\
			
			\hline
			\multirow{4}{*}{\rotatebox{90}{LSUI}} &FID$\downarrow$ & 98.49   & 100.5 & 34.51 & 38.90 & 158.99 & 45.06 & 28.56 & \underline{27.91} & \textbf{26.75}\\
			&LPIPS$\downarrow$ & 0.3962   & 0.3450 & 0.1432 & 0.1678 & 0.283 & 0.123 & \textbf{0.1028} & 0.1138 & \underline{0.1096}\\
			&PSNR$\uparrow$  & 15.43 & 18.24 & 20.86 & 19.73 & 22.63  &22.91& 24.16 & \textbf{27.65} &\underline{27.26}\\
			&SSIM$\uparrow$   & 0.7802 & 0.8465 & 0.8867 &0.8226  &0.9176 & 0.8902 & \underline{0.9322} & 0.8867 &\textbf{0.9437}\\
			\hline
			\multirow{2}{*}{\rotatebox{90}{U45}} &UIQM$\uparrow$ &   2.458  &  2.379  &  2.780  &2.957 &  2.856  & 3.104 & \underline{ 3.151 } & 3.086  & \textbf{ 3.181 }\\
			&UCIQE$\uparrow$ & 0.583   & 0.567 & 0.591 & 0.601 &  0.594 & 0.586 & 0.592 & \textbf{0.634}  & \underline{0.619}\\
			
			\bottomrule
		\end{tabular}
		
	\end{table*}
}

\begin{figure*}[t]
	\centering
	\includegraphics[width=1\linewidth]{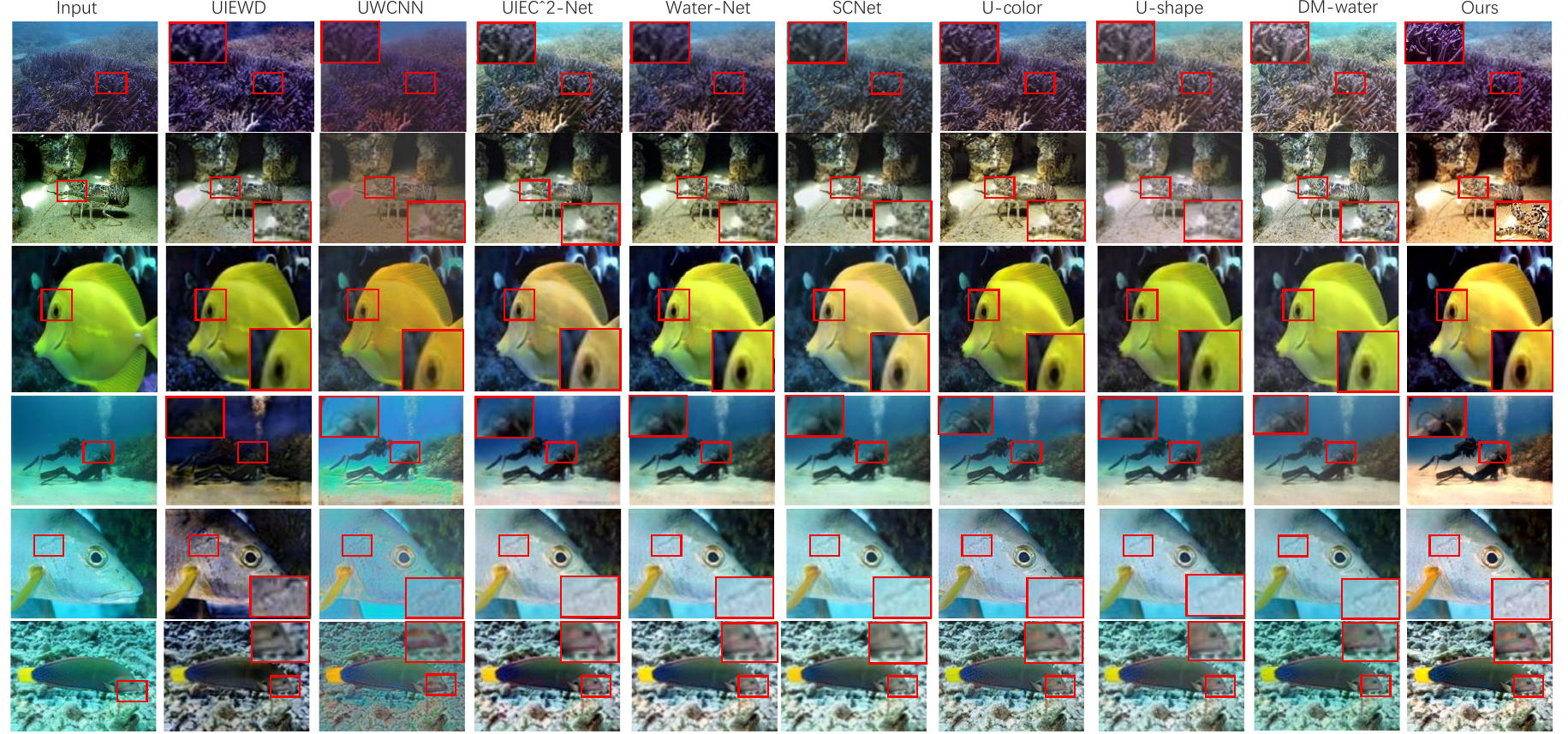}
	\vspace{-0.5cm}
	\caption{Qualitative comparison with other SOTA methods on real underwater images.} 
	\label{fig:2}
\end{figure*}

\subsection{Setup}
\noindent \textbf{Implementation details}. Our network, implemented using PyTorch 1.7, underwent training and testing on an NVIDIA GeForce RTX 3090 GPU. We employed the Adam optimizer with $\beta_{1}= 0.9$ and $\beta_{2}= 0.999$. The patch size was configured as 256 × 256, and the batch size was set to 2. The diffusion model's total time steps, denoted as T, were set to 1000, and the number of training iterations reached one million. The initial learning rate was established at 0.0001.
%Our network, implemented with PyTorch 1.7, was trained and tested on an NVIDIA Geforce RTX 3090 GPU. We use the Adam optimizer, $\beta_{1}= 0.9$ and $\beta_{2}= 0.999$. Our patch size was set to 256 × 256 and batch size to 2.  The total time steps of diffusion model T are set to 1000, and  The number of training iterations is one million. We set the initial learning rate to 0.0001. 
%EUVP dataset \cite{IslamXS20}, There are 12000 images for training and 515 images for evaluation in the EUVP dataset.

\noindent \textbf{Datasets}. We utilize the real-world UIEBD dataset \cite{PengZB23} and the LSUI dataset \cite{LiGRCHKT20} for training and evaluating our model. The UIEBD dataset comprises 890 underwater images with corresponding labels. Out of these, 700 images are allocated for training, and the remaining 190 are designated for testing. The LSUI dataset is randomly partitioned into 4500 images for training and 504 images for testing. In addition, to verify the generalization of WF-Diff, we use non-reference benchmarks U45 \cite{li2019fusion}, which contains 45 underwater images for testing.

%We employ the real-world UIEBD dataset \cite{PengZB23} and  the LSUI dataset \cite{LiGRCHKT20} to train and test our model. UIEBD contains 890 underwater images and corresponding labels. We use the 700 images for training and the rest for testing. The LSUI dataset was randomly divided as 4500 images for training and 504 images for testing. 

\noindent \textbf{Comparison methods}. We conduct a comparative analysis between WF-Diff and eight state-of-the-art (SOTA) UIE methods, namely UIECˆ2-Net \cite{WangGGY21}, Water-Net \cite{PengZB23}, UWCNN \cite{abs-1807-03528}, SCNet \cite{FuLWHD22},UIEWD \cite{MaO22}, U-color \cite{LiAHCGR21}, U-shape \cite{LiGRCHKT20}, and DM-water \cite{TangKI23}. To ensure a fair and rigorous comparison, we utilize the provided source codes from the respective authors and adhere strictly to the identical experimental settings across all evaluations.

%We compare WF-Diff with seven SOTA UIE methods including UIECˆ2-Net \cite{WangGGY21}, Water-Net \cite{PengZB23}, UWCNN \cite{abs-1807-03528}, SCNet \cite{FuLWHD22}, U-color \cite{LiAHCGR21}, U-shape \cite{LiGRCHKT20}, and DM-underwater \cite{TangKI23}.  For a fair comparison, we use the source codes provided by the authors and follow the same experimental settings.

\noindent \textbf{Evaluation Metrics}. We primarily utilize well-established full-reference image quality assessment metrics: PSNR and SSIM \cite{WangBSS04}. PSNR and SSIM offer quantitative comparisons of our method with other approaches at both pixel and structural levels. Higher PSNR and SSIM values signify superior quality of the generated images. Additionally, we incorporate the LPIPS and FID metrics for full-reference image evaluation. LPIPS \cite{ZhangIESW18} is a deep neural network-based image quality metric that assesses the perceptual similarity between an image and a reference image. FID \cite{HeuselRUNH17} measures the distance between the distributions of real and generated images. A lower LPIPS and FID score indicates a more effective UIE approach. For non-reference benchmarks U45, we introduce UIQM \cite{panetta2015human} and UCIQE \cite{yang2015underwater} to evaluate our method.

\subsection{Results and Comparisons}

Table 2 shows the quantitative results compared with different baselines
on UIEBD, LSUI and U45 datasets, including with UIECˆ2-Net, Water-Net, UIEWD, UWCNN, SCNet, U-color, U-shape, and DM-water. We mainly use PSNR, SSIM, LPIPS and FID as our quantitative indices for UIEBD and LSUI datasets, and UIQM and UCIQE for non-reference dataset U45. The results in Table 2 show that our algorithm outperforms state-of-the art methods obviously, and achieve state-of-the-art performance in terms of image quality evaluation metrics on
UIE task, which verifies the robustness of the proposed WF-DIff. To better validate the superiority of our methods, in
Figure 4, we show the visual results comparison with
state-of-the-art methods on real underwater images. The six examples are randomly selected on the UIEBD and LSUI datasets.
Our methods consistently generate natural and better
visual results on testing images, strongly proving that WF-Diff has good generalization performance for real-world applications.

\begin{table}[t]
	\caption{Ablation study with nerwork structure of WFI2-net on UIEBD dataset. }
	\centering
	
	\resizebox{0.85\linewidth}{!}{
		\begin{tabular}{c c c  c |c c}
			\specialrule{1.2pt}{0.2pt}{1pt}
			\multicolumn{2}{c}{WTB} &\multicolumn{2}{c}{SFFB} &  \multicolumn{2}{c}{UIEBD~} \\
			\cmidrule(lr){1-4}
			\cmidrule(lr){5-6}
			SA& CA & SDU & FDU & PSNR $\uparrow$ & SSIM$\uparrow$ \\
			\midrule
			\midrule
			$\times$  &$\checkmark$ & $\checkmark$ & $\checkmark$ & 20.94  & 0.8541\\
			$\checkmark$& $\times$ & $\checkmark$& $\checkmark$ & 20.82 &0.8473\\
			$\checkmark$ & $\checkmark$ & $\times$& $\checkmark$   & 21.11& 0.8586\\
			$\checkmark$ & $\checkmark$ & $\checkmark$ & $\times$& 20.23 & 0.8346\\
			\midrule
			$\checkmark$ & $\checkmark$ & $\checkmark$ &  $\checkmark$& \textbf{21.87} & \textbf{0.8622}\\
			\specialrule{1.2pt}{0.2pt}{1pt}
	\end{tabular}}
	\label{tab:loss}
	
\end{table}

\begin{table}[t]
	\caption{Ablation study with loss of WFI2-net on UIEBD dataset. }
	\centering
	
	\resizebox{0.85\linewidth}{!}{
		\begin{tabular}{c c c  c |c c}
			\specialrule{1.2pt}{0.2pt}{1pt}
			\multicolumn{4}{c}{Methods} &  \multicolumn{2}{c}{UIEBD~} \\
			\cmidrule(lr){1-4}
			\cmidrule(lr){5-6}
			$\mathcal{L}_{h}$ & $ \mathcal{L}_{rec} $ &  $ \mathcal{L}_{a} $& CFC & PSNR $\uparrow$ & SSIM$\uparrow$ \\
			\midrule
			\midrule
			$\times$  &$\checkmark$ & $\checkmark$ & $\checkmark$ & 20.46  & 0.8274\\
			$\checkmark$& $\times$ & $\checkmark$& $\checkmark$ & 20.65 &0.8399\\
			$\checkmark$ & $\checkmark$ & $\times$& $\checkmark$   & 19.81& 0.8213\\
			$\checkmark$ & $\checkmark$ & $\checkmark$ & $\times$& 20.97 & 0.8425\\
			\midrule
			$\checkmark$ & $\checkmark$ & $\checkmark$ &  $\checkmark$& \textbf{21.87} & \textbf{0.8622}\\
			\specialrule{1.2pt}{0.2pt}{1pt}
	\end{tabular}}
	\label{tab:loss}
	
\end{table}

\begin{table}
	\caption{Ablation study with FRDAM on UIEBD dataset. DM refers to  diffusion model (DM), RDM refers to residual DM. D-L refers to refining low-frequency component in wavelet space, and D-H refers to refining high-frequency components in wavelet space.}
	\centering
	\label{tab:freq}
	
	\begin{tabular}{c|ccccc|c}
		\specialrule{1.2pt}{0.2pt}{1pt}
		
		Method  &DM & RDM  & D-L  & D-H& CFC & PSNR$\uparrow$        \\\midrule
		
		A&$\checkmark$          & $\times$             & $\times$         &$\times$  & $\times$        & 20.86    \\
		B&$\times$  & $\checkmark$                     & $\times$         & $\times$    & $\times$     &22.37  \\

		C&$\times$          &  $\checkmark$                & $\checkmark$       & $\times$       & $\times$   & 22.32       \\
		D&$\times$         &  $\checkmark$              & $\times$      & $\checkmark$       & $\times$   & 22.58      \\
		
		E &$\times$   & $\checkmark$        &  $\checkmark$             &  $\checkmark$     & $\times$     &   23.44          \\
		\midrule
		F &$\times$   & $\checkmark$        &  $\checkmark$             &  $\checkmark$     & $\checkmark$     &   \textbf{23.86}          \\
		
		\specialrule{1.2pt}{0.2pt}{1pt}
		
	\end{tabular}
	
\end{table}
\begin{figure}[!t]
	\centering
	\includegraphics[width=\linewidth]{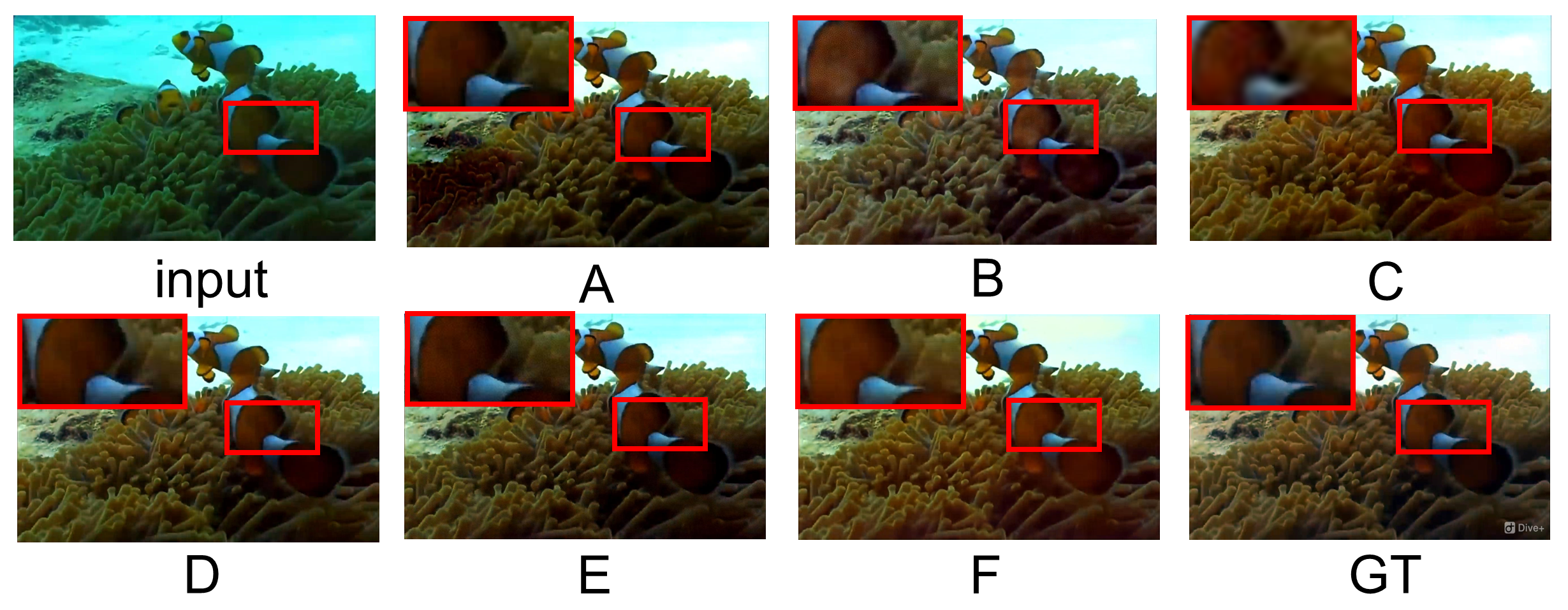}
	\vspace{-0.5cm}
	\caption{ Visual results of ablation study with FRDAM.  %From visual results, we find a similar phenomenon.%We swap the amplitude components of the underwater image and its corresponding ground-truth. The recombined result of the amplitude of label image and the phase of underwater image has similar visual appearance with label image. This phenomenon indicates the lightness degradation information of underwater images is mainly contained in the amplitude component. We further explore the properties of the amplitude components in Wavelet space. From visual results, we find a similar phenomenon.
	}
\end{figure}
\subsection{Ablation Study}
\textbf{Ablation study with WFI2-net.}
In order to evaluate the effectiveness of each part in WFI2-net, we conduct two ablation experiments with WFI2-net on the UIEBD dataset in Table 3 and 4. CFC is cross-frequency conditioner. Note that, we do not discuss the effect of our proposed FRDAM here. We regularly remove one component to each configuration at one time, and our strategy achieves the best performance by using all loss functions and blocks, proving that each part of WFI2-net is useful for UIE task.

\noindent
\textbf{Ablation study with FRDAM.}
In this section, we will discuss the effectiveness of FRDAM. Table 5 shows the quantitative results on the UIEBD dataset, and Figure 5 shows the visual results. Note that, Model A and B achieve diffusion process in the pixel level, and model C, D, E and E achieve diffusion process in wavelet space. Model A obtains relatively poor results in PSNR and generates images with color distortion or artifacts in Figure 5 due to the diversity of the sampling process. Model B does not yield fully satisfactory results, because it needs to adjustment both the high and low-frequency information in images, which limits their ability to focus on fine-grained information. Compared to model C, model D achieves better results, suggesting that the degradation information is mainly in the high-frequency information in frequency diffusion adjustment stage. Model F achieves the best performance, proving that our designed FRDAM is best for UIE task.
\vspace{-5pt}
\section{Conclusion}
\vspace{-5pt}
\label{sec:concl}

In this paper, we develop a novel UIE framework, namely WF-Diff. With full utilizing the frequency domain characteristics and diffusion model, WFI2-net can achieve enhancement and adjustment of frequency information. Our proposed FRDAM is a plug-and-play universal module to adjust the detail of the underwater images. WF-Diff shows SOTA performance on UIE task, and extensive ablation experiments can prove that each of our contributions is effective. As a result of employing two diffusion models, our approach doesn't confer an advantage in terms of inference speed. Hence, in the future, we will delve into methods to expedite the sampling process.

\clearpage
%%%%%%%%% REFERENCES
{\small
\balance
    \bibliographystyle{ieee_fullname}
    \bibliography{Reti-Diff}
}

\end{document}